# Context Agnostic Trajectory Prediction Based on λ-Architecture

Evangelos Psomakelis[1,2], Konstantinos Tserpes[1,2], Dimitris Zissis[3], Dimosthenis Anagnostopoulos[1] and Theodora Varvarigou[2]

[1]*Dept. of Informatics and Telematics, Harokopio University of Athens, Omirou 9, Tavros, Greece*
[2]*Dept. of Electrical and Computer Engineering, National Technical University of Athens, Iroon Polytecniou 9, Zografou, Greece*
[3] *Dept. of Product & Systems Engineering, University of the Aegean, Ermoupolis, Syros, Greece*
{vpsomak, tserpes}@hua.gr, dzissis@aegean.gr, dimosthe@hua.gr, dora@telecom.ntua.gr

## Abstract

Predicting the next position of movable objects has been a problem for at least the last three decades, referred to as 'trajectory prediction'. In our days, the vast amounts of data being continuously produced add the big data dimension to the trajectory prediction problem, which we are trying to tackle by creating a λ-Architecture based analytics platform. This platform performs both batch and stream analytics tasks and then combines them to perform analytical tasks that cannot be performed by analyzing any of these layers by itself. The biggest benefit of this platform is its context agnostic trait, which allows us to use it for any use case, as long as a time-stamped geo-location stream is provided. The experimental results presented prove that each part of the λ-Architecture performs well at certain targets, making a combination of these parts a necessity in order to improve the overall accuracy and performance of the platform.

## 1  Introduction

Trajectory prediction is a complex algorithmic challenge with practical implications in a variety of critical domains; For example, naval course plotting in dangerous, urban course plotting in avoidance of traffic, as a means of incident identification in naval or air routes, in big events planning as a means of congestion avoidance or resource allocation and many more. The challenge is that of predicting the position of a moving object, or person, after a pre-defined time period has elapsed. In order to predict this position, it is necessary to form a deep understanding of the object's usual moving patterns, even under the influence of external and internal factors that may affect its usual patterns. In most cases, as we will discuss in a following section, machine learning techniques are employed in order to create a prediction model based on both the usual behavior and all the factors affecting it, such as the wind, the natural obstacles, the human emotion and others.

In the past predicting an object's trajectory was largely tackled by dynamic, linear models obtained empirically. The prediction performance relied largely in the construction of equations that counted in the feasible points, the external uncertainty (e.g. wind, obstacles, etc) and external influence (e.g. using the controls of a vehicle to affect its speed). However, in our days, with an abundance of data becoming available to us, we can model these uncertainties based on data collected either offline in certain periods of time or in real time. For quite some time now big data analysis on offline (batch) data has been possible as we now have extremely powerful algorithms that extract information from large quantities of data in a matter of minutes, using distributed machine learning techniques and parallel computing.

On the other hand, real time analysis on streaming data is much more challenging. For a variety of real world applications, timeliness is a top priority. Time critical computing systems are systems in which the correctness of the system is dependent not only on the accuracy of the result produced, but also on the time in which it was computed; such systems include avionics and marine navigation systems, defense systems, command and control systems, robotics and an ever-increasing number of Internet of Things (IoT) applications. Trajectory analysis in the real world is more often based on streaming data under real time constraints than on batch data. For example, in our experiments we are using AIS (Automatic Identification System) data collected in a streaming fashion, with 1 second interval, from multiple radio receptors on the coastline. In other cases trajectory prediction needs to be performed on GPS data streamed from mobile phones or automobiles. The main difficulty when performing analytics on streaming data is that of performing incremental training of the models, while avoiding the always increasing usage of resources. Classical machine learning involves creating models by training an algorithm on a pre-defined dataset. In streaming data analytics that is not the case; we have firehoses of data, with records pouring in our endpoint continuously, so the need arises either for a model that is updated with each new record or to retrain a classical model using a window of records. This window can take the form of a sliding window, where each new record that enters the window pushes the oldest one out of it replacing it or a buffer of pre-defined size that is filled with the newest data at regular intervals. Of course, when using the window approach we are just disregarding all the historic knowledge so we cannot establish usual behavior patterns. On the other hand, when using the incremental learning algorithms we have to keep a watchful eye on the resource consumption of these algorithms or the summarization technique used in case it focuses on the wrong features.

In the related literature, many efforts have already been made at employing prediction models based on machine learning algorithms, both classic algorithms and customized ones. The problem with this type of solutions is that machine learning needs to know the prediction interval, how much time we want to elapse between the current position of the object and the predicted position. This knowledge needs to be pre-defined during the training of the model, a process that often takes hours to be completed if the dataset is too large. In real life application though, usually we do not have that knowledge before the need for prediction arises and when this need arises we need to have an answer in real time, we cannot wait for the machine learning algorithms to be retrained for the requested time-frame. Thus, the need for a platform that can answer in real time constraints or near real time about any time-frame arises.

Our approach is context agnostic, as long as the movement is limited to 2D space, as it is based on just five data points; the current latitude and longitude, the heading, the speed, the id of the object and the next latitude and longitude of the target object. It does not know nor care about the type of this object, the terrain it travels on or any other external or internal factor. Our aim is to model the behavior of the entity in a data-driven way, with minimal reliance on other sources of information. This creates a prediction platform that may be slightly less accurate than a specialized one but it can be applied with no effort to any application, regardless of its domain. It can predict the trajectories of ships in the ocean as well as the trajectories of visitors in a science fair or even the trajectory of data packs in a computer network. It is obvious that the benefits of a powerful, domain agnostic prediction platform are great, especially in applications that need to predict the trajectories of various types of objects at the same time. For example, a company could be organizing an automobile fair and be in need of a trajectory prediction platform for both the pedestrian visitors and the vehicles that run through the fair.

We employ a trajectory prediction platform, based on the lambda architecture. It creates a set of prediction models using machine learning on an AIS dataset provided by MarineTraffic for research purposes. More details about that dataset and AIS will be presented at a later section. We use this dataset as a real time emulated data stream, using the provided timestamps, at the same time storing each data pack in a backend database as it arrives to the stream endpoint. This way we practically have two data sources, an online data source (streaming) and an offline data source (batch). As we will present in a following section, the prediction error of each machine learning algorithm is relevant to its data source, some algorithms fare well with streaming data while others with a more complete but slow, batch data source. The source chosen in each case depends on the pre-defined prediction time-frame. We found that for shorter time-frames, for example the position of a vehicle after 2 seconds, the streaming data source has smaller error ratios while for larger time-frames, for example the position of a vehicle after 10 minutes, the batch data source fares better. Thus, we created streaming models for 2 minutes and 4 minutes and batch models for 10 and 20 minutes. Then we tried to answer queries for time-frames that are located in between these models, combining the predictions of the closest ones.

This work contributes to the field of trajectory prediction by providing a context agnostic platform for trajectory modelling, based on a lambda (λ) architecture. The trajectory modeling is done on batch and streaming data, using different algorithms on each occasion and then the predictions are combined forming one prediction that is closer to the truth than the predictions used to form it. Moreover, our platform supports queries about new time frames, that do not have a corresponding model, providing real time predictions by combining the nearest available models and extrapolating the position of the object at the requested time frame.

The rest of this paper is constructed as follows; in section 2 we present the definition of our research problem, the domain we deal with and the basic assumptions. In section 3 we provide an overview of the state of the art about trajectory prediction and streaming data analysis, in order to provide a basis for our work. In section 4 we present the preparatory steps we took before creating the proposed platform. Finally, in section 5 we provide a detailed description of the experiments we performed and an analysis of their results.

## 2   Problem definition

### 2.1  Domain

This research is focused on predicting the next position of a naval vessel after receiving its current AIS transmission, containing a plethora of information including, but not limited to, its geo-location, its heading, its speed and its destination. As "next position" we define the ships geo-location (latitude and longitude) after a pre-defined time interval has elapsed. This interval greatly affects the prediction as we will discuss in a following section.

AIS transmissions are globally recognized as an effective way of coordinating naval traffic and avoiding many accidents such as collisions or natural hazards. Every ship is transmitting an AIS record at regular intervals registering all information that other ships need in order to coordinate their own course. AIS uses a VHF transceiver as a transmission medium. In some cases a satellite is used, creating an S-AIS data stream [1].

The vessels contained in the dataset used are mostly naval vessels of different types. We could identify over 150 different types, of which dominant places hold the General Cargo, Bulk Carrier, Container Ship and Fishing. All vessels were moving inside the Mediterranean Sea and most of them on major naval pathways.

## 2.2 Assumptions and Features

Our work focuses on creating a context agnostic platform for trajectory prediction queries, using context irrelevant information such as the speed, geo-location and heading of the target object. Having that in mind we performed various experiments, that will be described in a following section, with two purposes; (a) to test which machine learning algorithms respond better to our dataset and the prediction intervals we tested and (b) to create a λ-architecture based platform that combines these prediction models in order to answer queries about prediction intervals without a pre-trained model on those exact intervals. We need to mention here that our approach aims to predict the next geo-location point of the target object, at a given time interval. This is a low level trajectory prediction problem, enhancing our aim to apply this platform on any context.

As mentioned, our models aim to predict the next position of an object after a time interval has elapsed. We chose five features for the models handling this prediction, based on the work of Valsamis et al [2]; (a) MMSI of ship, as an object id, (b) Speed Over Ground (SOG), thus including information about the rate of position change over time, (c) Heading, which is the facing direction of the object in degrees with 0/360 being the North and 180 being the South, (d) Geo-location of the object at current time and (e) Geo-location of the object after the time interval. As geo-location we refer to the latitude and longitude coordinates.

All these data were available in the AIS dataset that we used for the experiments with the exception of the geo-location after the time interval (called next geo-location). This was created as a pre-processing procedure that run through the dataset and compared pairs of AIS records for each ship and each timestamp. When we found two timestamps that matched the interval for the same MMSI we updated the next geo-location attribute of the earliest record, again based on the work presented in [2]. Our prediction target was the next geo-location, using the other four attributes as features in our prediction models.

## 3 Related Work

## 3.1 Trajectory prediction

For the last 3 decades at least, researchers and practitioners have been trying to address the challenge of accurately predicting an objects trajectory. We can see successful attempts of predicting the trajectory of objects as early as 1998. In [3], Liu et al, are tackling the problem of modeling the trajectories of mobile users in order to optimize the resource allocation of mobile cells, achieving acceptable results with their "hierarchical location prediction" (HLP) algorithm. HLP is based on a two layer mathematical model, the top layer was learning the usual movement patterns of users on a global, trajectory based, level. The second layer was trying to predict the next cell to be crossed by a user on a lower level, using kinematic data and a Kalman filtering process.

More recently, the naval traffic regulations made AIS emitters mandatory for most ship types, creating a vast data source of time-stamped trajectory data. Many researchers saw the opportunity of studying real life data at such quantities, advancing the field rapidly. One of these works, presented in [4] by Ristic et al, also uses a two layer prediction method. Firstly, they are trying to identify if the behavior described by the data is normal or anomalous, using an adaptive kernel density estimator, also known as the Parzen window method. Then all anomalous trajectories are discarded and they are predicting the next probable position of a naval vessel, following a normal moving pattern, by using a particle filtering method.

Ristic et al [4] are using the abnormality detection as a means of preprocessing the data. In later works [2], [5], [6], we see that pre-processing the data is of great importance, especially when machine learning is used for the creation of the models, either alone or in cooperation with the clustering algorithms. That importance has led to great improvements in this domain over time, creating more customized solutions, according to the research problem we are trying to solve in each case, the tools we are using and the features we decided to work with. Mao et al, in [7], present us with a nice example of advanced pre-processing. They created a measure of route complexity that is based on the cosine between the geo-location points that comprise the trajectory. This measure is then used to filter out any trajectories that are too simple or too complex for the training process. Following this process, they apply an algorithm that extrapolates missing geo-location points along the trajectory in order to have a more concrete training dataset. This dataset was fed to an ELM algorithm creating a prediction model for the next positions of a naval vessel.

As we have already mentioned, the data availability is already great and it is always increasing as more and more devices are intercepting AIS data packs from passing ships. This can quickly turn the trajectory prediction problem into a big data problem, with all its special dangers and problems. Nowadays, there are algorithms that can analyze huge quantities of data in about the same time as their predecessors did for just a small fraction of these data. These algorithms are based on parallel computing, providing solutions such as HBase analysis on cluster of computers running Hadoop. Wijaya et al [8], are trying to perform analysis on a big dataset of AIS records, predicting the normal trajectories of ships by identifying neighboring ships with similar behavior. If a vessel is following the same patterns as two or more other vessels its trajectory is considered normal and a prediction of its next positions can be made by studying the trajectory of the other vessels.

Each researcher starts the prediction efforts having in mind a specific research problem, often connected to a practical application. This leads to the creation of customized features for the clustering and classification algorithms, corresponding to the pre-defined research targets. For example, Fu et al in [6], counted the stop points and phase changes of a ship along its trajectory, using them to find patterns of movement, easing the clustering of trajectories. The negative in this process is that to discover these features, they used the SOG field of the AIS data, which Pallota et al, in [5] deems as unreliable, so the reliability of these features are in question too.

Some of the latest works are using DBSCAN clustering on AIS datasets in order to identify normal trajectories of naval vessels or points of interest and locate abnormal or anomalous activity [5], [6]. DBSCAN is excellent in point of interest identification because it can be parameterized in order to isolate certain behaviors that point to specific types of hotspots. For example, Pallota et al [5] suggest that setting the cluster at a Speed Over Ground (SOG) of less than 0.5 reveals the main ports in the dataset by clustering low speed movement vessels and docked ships in specific geo-locations. In addition, they mention that when this clustering process is combined with metadata about the ships that belong in each cluster we can draw useful conclusions about the target audience that a port has and the main business it deals with. Similarly, we could draw a number of other conclusions by looking at other metadata such as the size of ships which is correlated to the size of the port and the fuel that the port needs during vessel refueling, for example if a port serves mainly small fishing boats it does not consume much fuel as opposed to a port serving big cargo ships that cross the Pacific ocean.

## 3.2 Streaming data

As we have already mentioned, nowadays huge amounts of data are being created and uploaded on the internet each second. All these data are mostly useless unless we process them in a meaningful way, extracting information out of them. There are two ways of doing that, either online (streaming analysis) or offline (batch analysis). The offline analysis is about storing all the data in huge databases and then analyzing them with classical machine learning or clustering algorithms. It is obvious that due to the seer amount of data this process takes too much time.

To compensate, researchers are using parallel computing and distributed systems, often based on the MapReduce paradigm [9], in order to reduce the time needed for the algorithms to finish their analysis tasks. The other solution, that of the online analysis, is gaining more and more supporters in the last few years because it is providing us with acceptable results using much less resources. An interesting tool, used by a plethora of researchers, is called SAMOA [10], an open source platform that combines the benefits of both online and offline analysis on big data tasks. SAMOA is providing us with all the streaming algorithms that are using fast and efficient incremental techniques, either learning or clustering, while making it easy for the users to scale out, creating a distributed streaming analysis platform.

Apache is housing two projects closely connected to one another and highly relevant to stream analytics; Kafka and Samza [11]. Essentially, Kafka is a message processing mechanism that works as a data broker to the underlying analysis mechanisms that Samza provides. For now we will focus on Samza as this is more relevant to the stream analysis tasks we need to perform in our research. Samza is, in essence, expanding on the MapReduce paradigm, splitting the data into packages, processing in parallel each package and then combining the results in a stream. It is a framework consisting of three parts; stream reader, stream processor and stream writer. Both the stream reader and stream writer are connected to the Kafka framework, letting Samza deal with processing the data and not procuring them or serving the results to the user. The middle layer, the stream processor, is implemented by the users, according to their needs. What Samza provides is packaging of data, scalability using the Hadoop system, state management, metrics, monitoring and failure checks for processes. In addition if a process is detected as failed it is automatically restarted.

According to Carbone et al. [12], the drawback of SAMZA is that it does not provide state consistency guarantees or out-of-order processing (OOP). Apache Flink [12] on the other hand, is a workflow engine that enables the user to perform both batch and streaming analytics jobs in a scalable fashion. It defines three distinct notions of time for its processing tasks; a) event time, b) processing time and c) ingestion time. The first one, event time, is the time that each event actually occurs, the timestamp of the data in our case. The processing time is the time of the machine that processes the data, which may be out of sync with the event time, affecting their semantics or even their order. Finally, the ingestion time is the time that each event enters the processing platform, which tries to compensate between the semantics information that event time holds and the ease of execution that processing time offers. Flink can execute jobs based on any one of these notions of time, even providing OOP if needed by an application.

The solution that we based our platform on is called Lambda Architecture (λ-Architecture) by its creators, N. Marz and J. Warren [13]. They describe it as a robust and scalable system for big data management, a system that can answer big data queries with no delays while applying these queries on the whole database, not just fractions of it. The way they achieve it is by splitting the work into three layers; the batch, the serving and the speed layer. The batch layer is a simple database pre-processing mechanism that creates answers to arbitrary queries, using the whole dataset. This process creates a set of ready-to-serve answers to queries not yet asked. These answers are then pushed to the serving layer that indexes them and stands by for serving them when the actual queries arrive. The speed layer, on the other hand, is used to form fast, low accuracy answers to online queries, as they arrive. So the second job of the serving layer is to combine the pre-generated answers of the batch layer and the online answers of the speed layer.

## 4 Preparation

### 4.1 Machine Learning

#### 4.1.1 Stream Analysis

As we will see in section 5, the dataset we are using is pre-processed in order to be cleansed and in order for us to extracted a stream of feature vectors. The features contained in each vector are discussed in detail in section 2.2. After being extracted from the dataset, the feature vectors' stream is passed to the Streaming Analysis algorithm which is updated with each new batch of features received every second. This update process has to be faster than the speed of the stream, which in our case is one second, in order to prevent packet collisions. In our case it is, in both algorithms used. For the machine learning part we used the Weka java library, or more specifically, the LWL (Local Weighted Learning) [14] and IBk (Instance-Based learning with parameter k) [15] algorithms. The first one is based on logistic regression whereas the second one is based on the naïve Bayesian networks.

LWL and IBk, as provided by the Weka Java library, are especially made for streaming data so no training window was needed, they are updatable using one training vector at a time, and we could just input the vectors as we were getting them from the emulated stream. LWL though showed little promise in the current research task compared to the IBk algorithm in every prediction interval, so we decided to discontinue the experiments with LWL and continue with just the IBk. In Table 1 we can see the error statistics for the two algorithms tested. As discussed in section 5.3, we chose to rely on the distance in nautical miles between the predicted geo-location and the actual geo-location as an error measure, which is also the measure used in Table 1.

The stream processing performed with an acceptable prediction accuracy in most nautical use cases (average error of 1.21 nautical miles in 2' predictions, 1.30 in 4' predictions and 1.21 in 10' predictions). The drawback is that the prediction accuracy is highly erratic, ranging between 0 – 787 nautical miles in 2' predictions, 0 – 604 in 4' predictions and 0 – 861 in 10' predictions with an always escalating error deviation. This erratic behavior is attributed to the ability of streaming algorithm to forget as time passes. So, for each new ship that reappears in the dataset we probably would have little to no knowledge of its general behavior or of behavior that similar ships exhibit, especially in longer prediction intervals. That leads the model to perform wild guesses until it has more recent data concerning this ship's behavior.

| Interval | Algorithm | Minimum | Average | Maximum | Median | S Deviation |
|---|---|---|---|---|---|---|
| 2' | LWL | 0,53870085 | 121,33670 | 500,252985 | 117,053900 | 6177,74430 |
| 2' | IBk | 0 | 1,2148487 | 786,990532 | 0,51682156 | 2802,39851 |
| 4' | LWL | 1,67044309 | 118,45057 | 500,926447 | 113,201232 | 6398,79524 |
| 4' | IBk | 0 | 1,3034549 | 603,896671 | 0,55023744 | 2695,55598 |
| 10' | LWL | 0,79862566 | 121,55521 | 475,866859 | 117,224335 | 6374,5113 |
| 10' | IBk | 0 | 1,2093172 | 861,055665 | 0,50220356 | 3255,74503 |

Table 1: Error statistics for streaming data algorithms.

As already mentioned, we had to discontinue the LWL experiments due to the fact that it underperformed the IBk algorithm in each tested scenario and almost all metrics measured, so from now on we will be using only the IBk algorithm for the streaming analysis tasks.

### 4.1.2 Batch Analysis

The Batch Analysis models are created again using the Weka java library. For the batch analysis we tested again the LWL and IBk algorithms in order to have a measure between the streaming and batch jobs. In later experiments we also tested five traditional machine learning algorithms. The first tested algorithm was Support Vector Machines [16]–[18] which uses regression functions in order to perform classifications. Following that, we tested two more function based algorithms, the Least Median Square [19] and Isotonic Regression [20] which use similar methods to perform their classifications. Moving on, we tested a random decision tree algorithm called M5L [21], [22] which creates decision trees based on probabilities. Finally, we tested an artificial neural network algorithm that we knew as effective in similar cases [2], called Multilayer Perceptrons [23].

All these algorithms were also provided by Weka and we tested them by applying them on the whole dataset collected up to the time of training (or retraining). Every time the algorithm was trained we measured the training time needed and set the next retraining for twice that time, in order to have the model trained as frequently as possible while avoiding a constantly in-training model that cannot give predictions. A direct conclusion from this set of experiments is that batch analysis is less erratic in its predictions, having smaller deviations, but it has less accuracy than the stream analysis, at least in most of the tested cases. In some of the experiments though, we have to mention that the streaming analysis was expanding its error range drastically as the interval grew whereas the batch layer was keeping it relatively steady. Regarding the comparison between the algorithms, as we can see from Table 2, which gives us the error statistics (using the distance in nautical miles as an error measurement again) for all algorithms tested, we can conclude that M5P and Isotonic Regression are the most reliable algorithms having the smaller Maximum and Average error. From these two, M5P seems to perform slightly better but Isotonic Regression is faster and demands less resources than M5P in a degree that makes it a better choice for our experiments. So, for the λ-architecture platform that will be described in a later section, we picked Isotonic Regression as a batch layer machine learning algorithm.

| Interval | Algorithm | Minimum | Average | Maximum | Median | S Deviation |
|---|---|---|---|---|---|---|
| 2' | SVM | 1,44461462 | 300,1624 | 1096,96345 | 315,822543 | 42380,9918 |
| 2' | MLP | 0,13429169 | 336,7958 | 2968,95352 | 310,298586 | 63944,0811 |
| 2' | M5P | 0,00068342 | **0,566267** | **337,612926** | **0,38907168** | **1040,45881** |
| 2' | LMS | 0,00131598 | 124,0126 | 6643,70734 | 0,39230169 | 178789,704 |
|   | Isotonic Regression | **0,00056579** | 58,61143 | 743,868292 | 0,54076380 | 36311,0018 |
| 4' | SVM | 3,21415313 | 306,7665 | 1856,13242 | 312,962546 | 45216,1082 |
|   | MLP | 0,07404567 | 405,1494 | 3056,24368 | 332,941886 | 108466,858 |
|   | M5P | **0,00039011** | **0,856354** | **732,123543** | 0,40631810 | **2066,06334** |
|   | LMS | 0,00102449 | 14,24914 | 6325,26726 | **0,39654813** | 66337,3250 |
|   | Isotonic Regression | 0,00168341 | 38,70157 | 738,499153 | 0,53523226 | 32510,9813 |
| 10' | SVM | 3,46780724 | 323,8435 | 793,606678 | 302,704617 | 58329,0753 |
|   | MLP | 0,22116321 | 595,2997 | 3036,39489 | 556,106697 | 103171,364 |
|   | M5P | 0,00260994 | 4,302324 | **640,586518** | 0,91907636 | **6681,63955** |
|   | LMS | **0,00049664** | **2,525573** | 4123,97973 | **0,54888524** | 12318,0497 |
|   | Isotonic Regression | 0,00060175 | 11,16249 | 736,424873 | 0,95068687 | 16915,3328 |

Table 2: Error statistics for batch analysis.

## 4.2 λ-Architecture

λ-Architecture refers to a specific architecture type that tries to combine two streams of information into one in order to improve either the integrity or the availability of data. In our case, we are using it in order to combine predictions from heavyweight models, trained using all available data (batch layer), and lightweight models, trained using the streaming data with small windows or summarization techniques (streaming layer). This is the case because, as mentioned earlier, we experimentally found that each model type has its own strengths and weaknesses. For example, the streaming layer performs well when we try to predict the position of object in short time intervals because it is always updated, knowing the latest position of the object and the course it is following, whereas the batch layer is more accurate for longer time intervals, having a more complete picture of the usual moving patterns of the object.

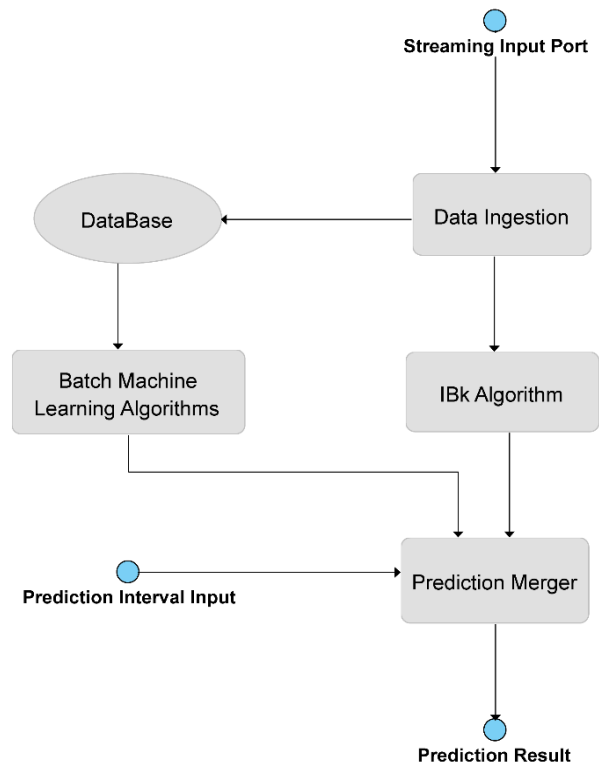

Figure 1: λ-Architecture general diagram

In Figure 1 we can see a general schema of the λ-Architecture platform created. As discussed in section 3.2, the λ-Architecture traditionally has 3 layers; batch, speed and serving. In our case we implement (a) the batch layer using a combination of the data ingestion, the database and the batch machine learning sub-components, (b) the speed layer using the IBk Algorithm and (c) the serving layer using the prediction merger. In detail, all AIS data are fed to the platform through a streaming input port as an emulated stream of realistic time scale. Then the data is passing through an ingestion mechanism that firstly stores each new data point in a database, in order to create the training set for the batch layer and at the same time it forwards the data, after performing simple data pre-processing tasks, to the streaming layer analysis algorithm. For the streaming analysis we picked the IBk algorithm due to its performance during the experiments mentioned in a previous section. The batch machine learning algorithm used depends on the prediction intervals set. Then both the batch and streaming layers are combined in the prediction merger component which reads the target prediction interval from the prediction interval input port and then combines the two predictions. More information about the merging process will be presented in a following section. After that the result is forwarded to the prediction result port. Of course, both the streaming and the batch algorithm components are comprised of a number of sub-components, one for each prediction interval we chose to train models for. In our case, we have models for 2 and 4 minutes in the streaming component and for 10 and 20 minutes in the batch component.

## 5 Experiment

### 5.1 Dataset

The experiments related to this research were contacted using a static csv file of 16.708.373 AIS records, containing, among others, information about the time stamped geo-location, speed, heading and ship type of ships travelling in the Mediterranean sea during the period 00:00:00 2015-08-01 to 02:43:39 2015-08-23. These records were provided by MarineTraffic for research purposes. In the following figure (Figure 2) we see a geo-location visualization of just 100.328 of the records using the Weka platform. As we see, there are some obvious pathways that ships prefer to move on as well as outliers and "dead zones" that present little to no activity. The colors describe the ship type that each AIS record corresponds to.

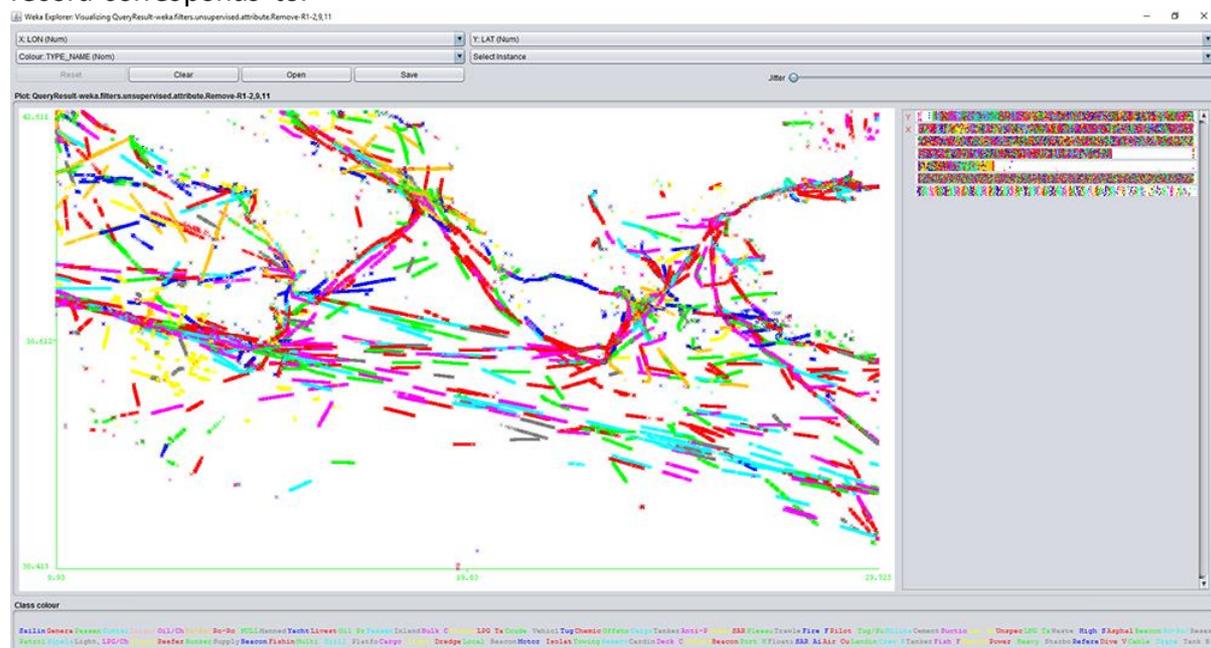

Figure 2: Geo-location visualization of a small data sample.

The original AIS stream used to create the CSV had a one second refresh interval, which means that every second a new batch of AIS records were registered. This time interval was followed in the experiments (using the provided time stamps), recreating the stream in real time. This allows us to study if the algorithms can respond fast enough with enough confidence.

The features used in the machine learning tools were extracted by preprocessing the emulated AIS stream. The prediction targets were the "Next Latitude" and "Next Longitude" of the ship in 2', 4' and 10'. The features used in order to predict this were the following: {"current latitude", "current longitude", "speed", "heading", "MMSI"}. For each record coming into the stream, if it corresponded to the pre-set timeframe, there was a 60% chance of it being part of the training set, 30% chance of it being part of the testing set for each algorithm and 10% chance of it being part of the testing set for the prediction platform. This extraction creates a feature vectors' stream for Streaming Analysis and a new record for each vector in the database, which is used in the Batch Analysis layer.

No other preprocess was applied to the data, in order to emulate as closely as possible a real data stream scenario, and to avoid domain specific data cleansing techniques. As discussed, one of our main targets is creating a domain agnostic technique of trajectory prediction.

## 5.2 Setup

As already mentioned, in our experimental process, we tried to predict the next position of a naval vessel after a pre-set time interval has elapsed and while knowing the current position of the vessel. The experiments were run in two phases. During the first phase, we tried to identify the most fitting machine learning algorithms for our specific research problem. We tested a variety of algorithms, all provided by the Weka java library [24]. In detail, we tested two streaming algorithms and seven batch algorithms, creating three different models in each case, one for a two minute prediction interval, one for a four minute interval and one for ten minute interval, as described in Section 3.2.

During the second phase, we tried to combine these algorithms in order to answer queries about prediction intervals that fell in between the created models. In detail, we tested queries about 185.94'', 360'' and 830'' intervals. In order to answer these queries we picked the best algorithm for each one of four arbitrarily chosen intervals (120'', 240'', 600'' and 1200'') and we created a model for each one. As we will discuss in the results section of this paper, we used streaming analysis models for the two lowest intervals and batch analysis for the two largest for prediction stability reasons. Then we predicted the requested vessel's position using the closest models to the actual query and applied a linear transformation to these predicted positions in order to estimate its final predicted position.

The process was simple, we received the prediction interval requested each time and found the closest model with less interval and the closest with more interval so that the requested interval fell somewhere in between them. If the requested interval was too small or too large only one model was used, the one with the smallest or largest interval respectively.

After getting these predictions we performed a linear transformation using function 1, in order to calculate the exact position of the ship between the two predicted positions.

$$fp(fv_i) = p_{i1} + \frac{int_1 * (p_{i2} - p_{i1})}{int_2}$$

*Function 1: Linear transformation.*

where:
- fp = final prediction
- $fv_i$ = feature vector i
- $p_{i1}$ = first known/predicted geo-point of vector i (the latitude or longitude of vector i if the prediction interval is too small or the predicted latitude or longitude from the lowest interval model used if prediction interval is greater than at least one model's)
- $p_{i2}$ = last known/predicted geo-point of vector i (the latitude or longitude of vector i if the prediction interval is too big or the predicted latitude or longitude from the largest interval model used if prediction interval is less than at least one model's)
- $int_1$ = the requested prediction interval in milliseconds.
- $Int_2$ = the prediction interval of the largest model used in milliseconds.

The resulting prediction was compared to the real position of the ship in that time in order to find out the prediction error and then this prediction error was compared to the error we would have if we just used the closest model available with no linear transformation, in order to estimate the improvement or deterioration that linear transformation causes to the predictions.

## 5.3 Results

The nature of the prediction target, which in our case was the future geo-location of a vessel, is a continuous value and as such it is almost impossible to have an exact prediction. That is why we chose to rely on "error in nautical miles" as our evaluation factor instead of the traditional mean squared error and kappa values. Of course, we applied some statistical formulas on this basic factor, estimating the average error, the standard deviation, the median error and the maximum/minimum errors for every tested algorithm.

In the rest of this section we will present the most important results of the experiments we performed. For each prediction interval, we will firstly present how the average error distance and the average error in latitude and longitude of the prediction algorithms is changing over time in order to have a first look in the learning progress. Then we will present the actual results of the prediction, comparing the algorithms used and the combinations of those algorithms in a λ platform.

### 5.3.1 Training progress

As we can see in Figures 3, 4 and 5, the learning curve is improving steadily over time, producing better results as more data are pouring into the model, in all three of the tested time intervals. This experiment provides proof that the over-fitting threshold that exists in every machine learning algorithm was not reached by the time these specific experiments were concluded. As such, we estimate that the results presented bellow can be further optimized with a larger dataset, providing more training data to the models.

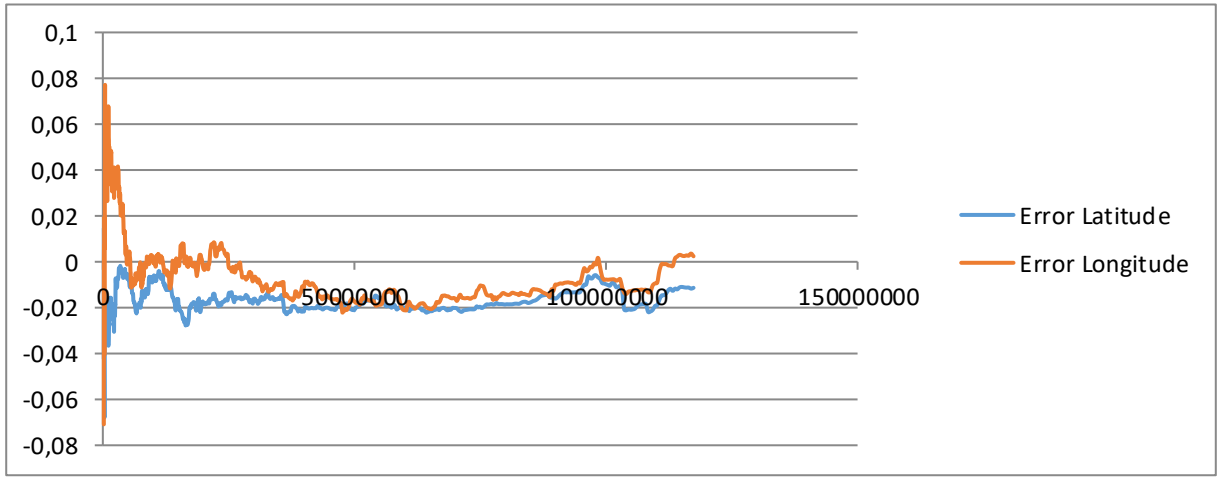
Figure 3: Latitude and Longitude error over time for 120 sec interval.

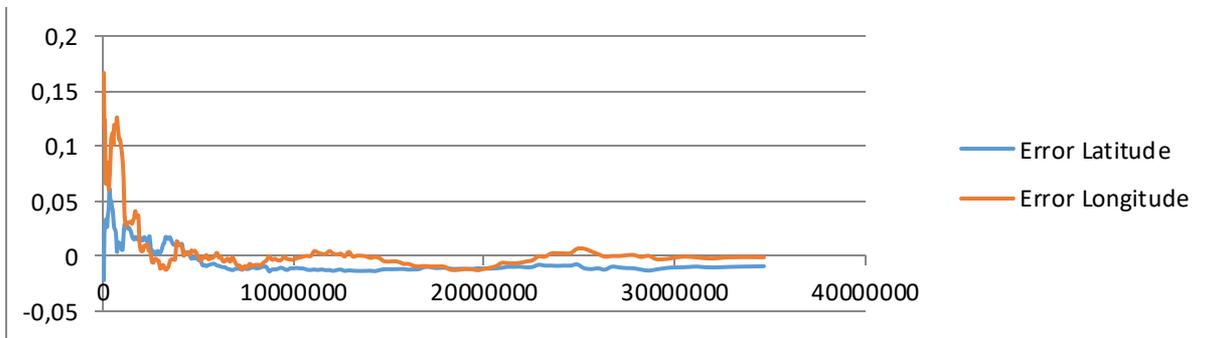
Figure 4: Latitude and Longitude error over time for 240 sec interval.

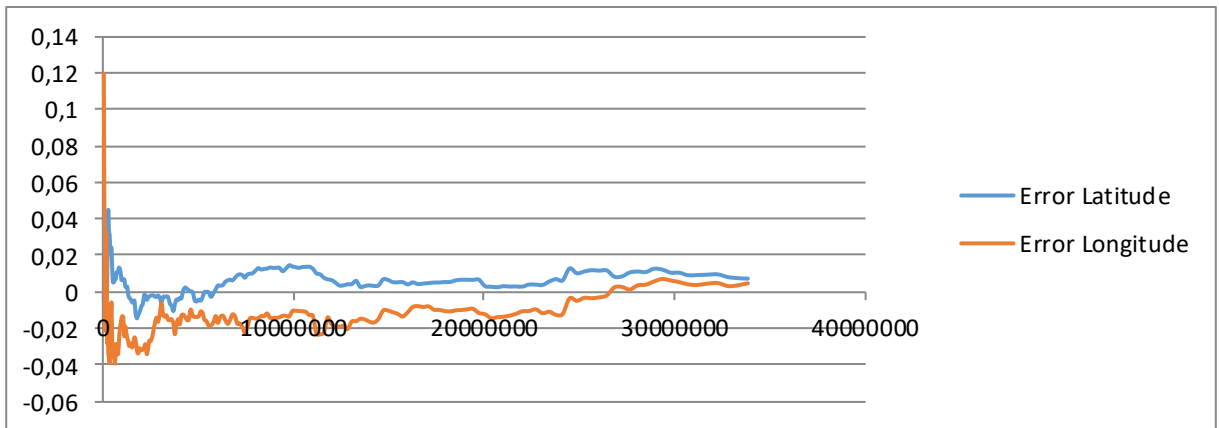
Figure 5: Latitude and Longitude error over time for 600 sec interval.

### 5.3.2 λ-Architecture Results

As discussed, our prediction platform, which we will call "λ-platform" from now on, is based on the λ-architecture, using the IBk algorithm in order to perform analysis on streaming data and the Isotonic Regression algorithm in order to perform analysis on a backend database, referred to as batch data. These two analysis tasks are then combined in order to answer a query that neither of these tasks could answer by itself.

We run three groups of experiments for three different prediction interval queries and the results are presented in Table 3. As mentioned in Section 5.2, we compare 2 types of results, the "base" one that just takes the closest available model to the interval requested by the query and the "λ-platform" one which applies the linear transformation in order to combine the closest available models of the platform, providing a merged prediction.

| Interval | Label | Base | λ-platform |
|---|---|---|---|
| 185940 | Minimum | 0 | 0 |
| | Maximum | 368,3375 | 396,7692 |
| | Average | 2,120542 | 5,223758 |
| | Median | 0,679229 | 3,483614 |
| | Deviation | 969,2249 | 1142,743 |
| 360000 | Minimum | 0 | 0 |
| | Maximum | 320,2301 | 300,7792 |
| | Average | 2,5672 | 2,583114 |
| | Median | 1,055959 | 1,074027 |
| | Deviation | 1171,176 | 1072,595 |
| 830000 | Minimum | 0 | 0 |
| | Maximum | 389,6276 | 389,6276 |
| | Average | 5,69674 | 5,519548 |
| | Median | 2,050329 | 2,01818 |
| | Deviation | 2228,45 | 2158,334 |

Table 3: λ-platform error statistics.

We can see that the platform did not behave well for the first group of tests which is the most difficult one due to the nature of the prediction interval chosen. The 185940 milliseconds interval is not only foreign to the pre-trained models but also foreign to the native 1 second step of the source data. This creates a problem for the machine learning algorithms that is totally out of their training data, causing them to behave erratically and actually worsen the predictions, adding to their error rates.

In the 360000 milliseconds (6 minutes) group, we see that the average and median error rates are actually slightly worse, even if the difference is negligible in practical use, but the improvements in standard deviation and maximum error shows us that the platform has achieved to balance the erratic behavior of streaming analysis algorithms and create a more reliable prediction model. Finally, looking at the 830000 milliseconds (13 minutes and 50 seconds) group, we are starting to see improvements in most metrics, having less average and median error rates as well as a smaller standard deviation.

We have to note here that in the first group the platform combined the predictions of two streaming layer algorithms (2' and 4'), in the second group the platform combined one streaming and one batch layer algorithms (4' and 10') and in the final group the platform combined two batch layer algorithms (10' and 20'). This is the case because, as discussed earlier, the platform searches for the nearest available pre-trained models in order to merge them, using a linear function, and create the final prediction.

### Conclusions

A direct conclusion from the experiments conducted is that batch analysis is less erratic in its predictions, having smaller deviations, but it has less accuracy than the stream analysis, at least in most of the tested cases. In some of the experiments though, we have to mention that the streaming analysis was expanding its error range drastically as the interval grew whereas the batch layer was keeping it relatively steady.

In most tested cases though, the Streaming Analytics outperformed the Batch Analytics making the combinations irrelevant if the use case demands high accuracy and is able to ignore a small percentage of the predictions that are way off, as we can just rely on the Streaming Analytics. In a use case that prediction stability is more important than accuracy though this combination can prove useful. The steady predictions of the batch layer as the prediction interval grows could also prove useful if the prediction intervals are too large for the stream analytics to be effective but this has to be tested in later experiments in order to prove that this behavior continues as expected in greater intervals, which is one of the main targets for our future work on this platform.

In our experiments we tested three intervals one that is totally out of the one second interval of the original one second stream input rate of the AIS data (185.94''), one that is between the last stream analysis model and the first batch analysis model (360'') and one that is between two batch analysis models (830''). The results show a steady improvement for the last two tests. On the other hand, as expected for the first test, the results show a degradation of the accuracy with the usage of the linear transformation. This may be attributed either to the totally foreign nature of the interval (it is between the 185th and 186th seconds) or the more erratic nature of the stream analysis algorithms that could be affected by the linear transformation negatively. The second target for our future work is to investigate these possibilities.

This experimental platform will serve as a test bench, allowing us to test different algorithms, both online and batch, in different use cases in order to find optimal combinations between the domains, the features and the algorithms, with the optimization targets being the accuracy and the resource efficiency. The context agnostic nature of the platform allows us to do that, also allowing us to use the platform in a variety of other experiments.

## Vitae

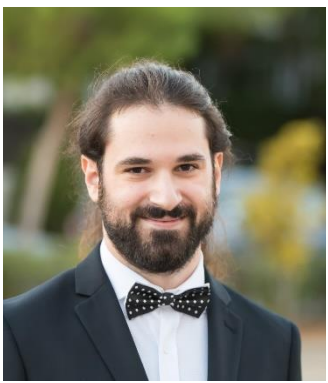

**Evangelos Psomakelis** is a PhD candidate at the department of Telematics and Informatics of Harokopio University of Athens and a Research Associate at the Distributed, Knowledge and Media Systems (DKMS) of the National Technical University of Athens. He is a graduate of Telematics and Informatics of the Harokopio University of Athens, holding a Master in Computer Science degree from the same university. His main research interests include machine learning, data mining and sentiment analysis. He was involved in the European Union founded project "Consensus" and he is currently involved in the cooperative European and Korean project "BASMATI".

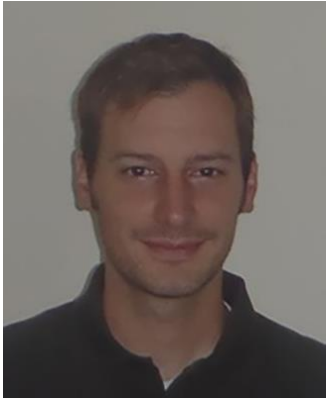

**Konstantinos Tserpes** is an Assistant Professor at the department of Telematics and Informatics of Harokopio University of Athens and a Research Associate at the Distributed, Knowledge and Media Systems (DKMS) of the National Technical University of Athens. He is a graduate of the Computer Engineering and Informatics department of the University of Patras. He holds a PhD in the area of Service Oriented Architectures with a focus on quality aspects from the school of Electrical and Computer Engineering of NTUA and a Master's degree from the same School. His research interests revolve around distributed and heterogeneous systems and software and service engineering.

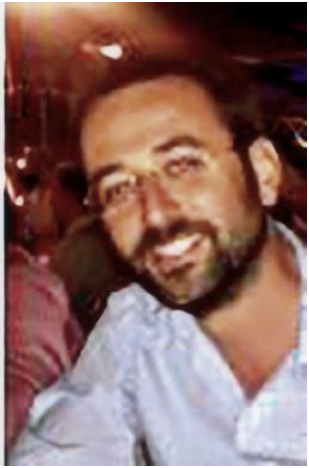

**Dimitris Zissis** is an Associate Professor at the University of the Aegean Department of Product & Systems Engineering. His research interests and areas of expertise include several aspects of architecting and developing complex Information Systems, including distributed and cloud based big data deployments. His published scientific work includes more than 50 publications, while he has participated in several European research projects. He is an editor for Future Generation Computer Systems, published by Elsevier while in the past he has guest edited and chaired numerous special issues and conferences. He is a member of the IEEE, the IEEE Computer Society and the IEEE Intelligent Transportation Systems Society.

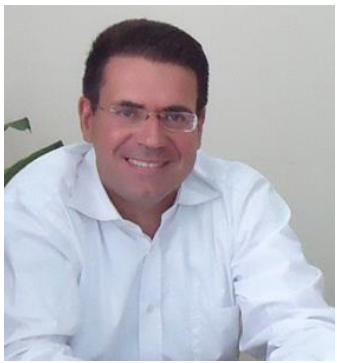

**Dimosthenis Anagnostopoulos** is a Professor in the Department of Informatics and Telematics in the area of Information Systems and Simulation. He served as Rector of Harokopio University (9/2011 – 1/2016). Since 1/2016, he is the Dean of Faculty of Digital Technology. He was elected Chair of the Greek Rectors' Conference (1/2014-6/2014). He is a visiting Professor at Sussex University, UK. He is an Associate Editor of Requirements Engineering (Springer). He served as the National Representative of Greece for ICT in Horizon 2020 (2014-2015). His research interests include Information Systems, eGovernment, Semantic Web and Web Services, Modelling and Simulation, Business Process Modelling.

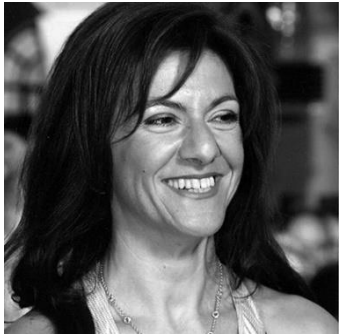

**Theodora Varvarigou** is a professor at the School of Electrical and Computer Engineering of the National Technical University of Athens (NTUA). She received the B. Tech degree in Electrical Engineering from NTUA in 1988, the MS degrees in Electrical Engineering (1989) and in Computer Science (1991) from Stanford University, California. She received her Ph.D. degree from Stanford University as well in 1991. She has worked as a researcher at AT&T Bell Labs, USA and as an Assistant Professor at the Technical University of Crete, Chania, Greece. Prof. Varvarigou has experience in technologies, such as Cloud computing, semantic web, social networking technologies etc.